\documentclass[10pt,twocolumn,letterpaper]{article}

\usepackage{wacv}              
\usepackage{booktabs,tabularx}
\usepackage{adjustbox}
\usepackage{multirow}
\usepackage{makecell} 
\usepackage{soul}

\usepackage{float}
\usepackage{graphicx}      
\usepackage{amssymb}
\usepackage{amsmath}
\usepackage{threeparttable}
\usepackage{tablefootnote}
\usepackage{pifont}
\usepackage{booktabs}
\newcolumntype{Y}{>{\raggedleft\arraybackslash}X}
\definecolor{wacvblue}{rgb}{0.21,0.49,0.74}
\usepackage[pagebackref,breaklinks,colorlinks,allcolors=wacvblue]{hyperref}

\def\wacvPaperID{*****} 
\def\confName{WACV}
\def\confYear{2026}

\title{HRDX: A Large-Scale Vector HD-Map Dataset}

\author{
Sahith Reddy Chada \hspace{5em}
Isht Dwivedi \hspace{5em}
Nirav Savaliya \\
Honda Research Institute US, San Jose, CA. \\
{\tt\small \{sahith\_chada, idwivedi, nsavaliya\}@honda-ri.com}
}
\begin{document}

\twocolumn[{%
\maketitle
\begin{center}
    \vspace{-0.8em}
    \includegraphics[width=\textwidth]{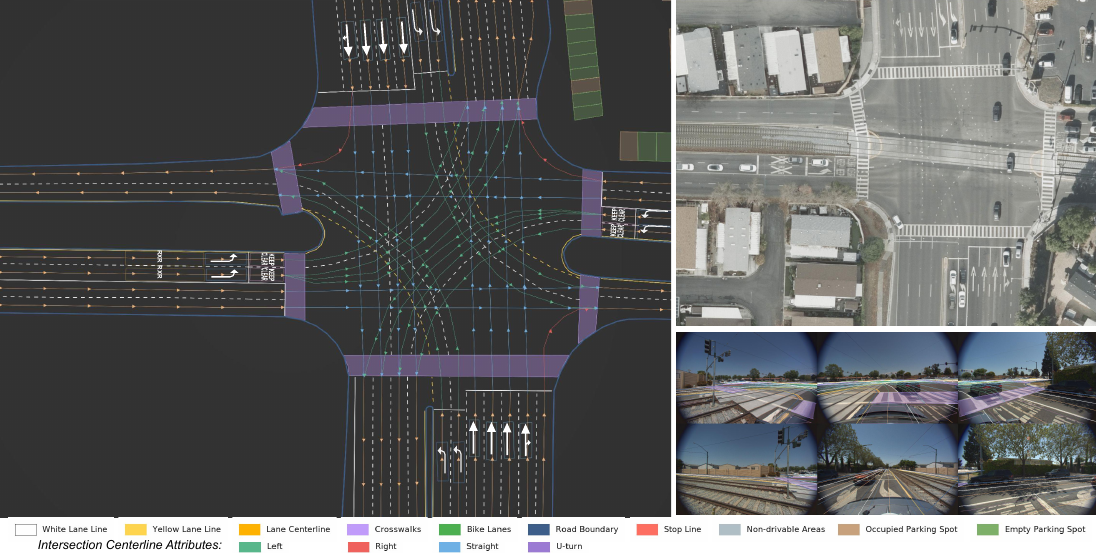}%
    \vspace{-0.5em}
    \captionsetup{hypcap=false}
    \captionof{figure}{Sample scene from the HRDX dataset showing vectorized map elements (left), aligned aerial imagery (top right), and camera views with projected map annotations (bottom right).}%
    \label{fig:teaser}
    \vspace{1em}
\end{center}
}]
\begin{abstract}
Reliable autonomous driving requires vectorized HD maps that are geometrically accurate, semantically rich, and scalable to long-horizon driving. However, existing public HD map datasets are limited in scale, provide sparse semantic attributes, and lack modalities such as aerial imagery that could enable new research directions. We present HRDX, a large-scale dataset for vector HD-map construction, spanning about 40 hours (1,400 km) of minimally overlapping drives, which is several times larger than prior public HD map datasets. Data is captured using six synchronized surround cameras, a 128-beam LiDAR, and centimeter-level RTK GNSS/IMU, and is further complemented by precisely aligned aerial orthoimagery. Annotations cover 10 vector map classes, complemented with over 20 semantic and topological attributes. To evaluate this richer ontology, we introduce the Composite Score (CS) to jointly assess geometric fidelity and attribute correctness. Benchmark experiments show that HRDX's scale improves online vector-map construction, and that aligned aerial imagery provides a useful structural prior: using aerial imagery at training and/or inference improves geometric map quality, while aerial-augmented teachers can transfer part of this benefit to camera-only students without increasing inference-time sensor requirements. HRDX is intended to support reproducible research on large-scale HD-map learning, multimodal BEV fusion, and training-time privileged information. HRDX dataset and benchmarks are available at \url{https://github.com/honda-research-institute/HRDX}.
\end{abstract}
   
\section{Introduction}
\label{introduction}
Autonomous driving systems require a precise understanding of static road elements such as lane lines and road boundaries. Inferring vectorized map elements from onboard sensors in real time has emerged as a scalable alternative to pre-built HD maps, which require specialized fleets, extensive annotation, and continuous maintenance~\cite{li2022hdmapnet,liu2023vectormapnet,liao2025maptrv2,chen2024maptracker}. Rapid progress on model architectures has driven substantial accuracy gains, yet public benchmarks have not kept pace. Existing HD map datasets are limited in driving coverage (typically under 10 hours), annotate few semantic attributes beyond geometry, and lack modalities such as aerial imagery that could unlock new research directions. As a result, it remains unclear whether current performance plateaus reflect fundamental model limitations or simply insufficient data scale and annotation richness.

We introduce HRDX, a large-scale dataset designed to close this gap. Collected across diverse urban, suburban, and freeway environments in the San Francisco Bay Area, HRDX spans over 40 hours and 1{,}400\,km of minimally overlapping drives, approximately \emph{six times} larger than the next largest public HD map dataset. The platform captures data from six synchronized surround cameras, a 128-beam LiDAR, and centimeter-level RTK GNSS/IMU, and provides aerial imagery. Annotations cover 10 vector map classes and over 20 attributes (Table~\ref{tab:annotation-ontology}), including lane centerlines, lane lines with color and style, stop lines, crosswalks, bike lanes, parking spots, non-drivable areas, and a rich taxonomy of road-surface text markings (arrows, STOP, YIELD, keep-clear zones). Some of these classes and attributes, which encode regulatory and behavioral semantics that downstream planners must observe, are entirely absent from all prior public HD map datasets.

A distinctive feature of HRDX is the inclusion of precisely aligned aerial orthoimagery at up to 8\,cm/pixel resolution. While high-resolution overhead imagery has been widely used in remote sensing for road-graph extraction~\cite{bastani2018roadtracer,li2019polymapper,vanetten2018spacenet}, and recent work has explored fusing satellite tiles with onboard sensors for HD mapping~\cite{lengerer2025aid4ad,gao2024complementing,zhao2024opensatmap}, no existing public HD map dataset provides aerial imagery aligned to vehicle trajectories at this resolution. Some prior works augment nuScenes with aerial tiles obtained via Google Maps APIs for experimental use; however, systematic downloading or redistribution of such imagery is prohibited under Google's Terms of Service, limiting reproducibility and public release. In contrast, HRDX sources its aerial imagery from publicly available county orthoimagery databases~\cite{sf_aerial_2024,sanmateo_imagery_2022,sanjose_arcgis_nodate}, enabling unrestricted redistribution and reproducible research on cross-view fusion, aerial-supervised map learning, and training-time privileged information.

We demonstrate the utility of aerial imagery by incorporating it during training, either as a direct multimodal input or as a training-only signal via knowledge distillation from an aerial-augmented teacher to a camera-only student and show that both settings yield consistent improvements in geometric fidelity and semantic attribute prediction, illustrating how HRDX's aligned overhead imagery can serve as a practical, scalable structural prior for deployment-ready mappers. To standardize evaluation on this richer annotation space, we propose Composite Score (CS), a metric that jointly assesses geometric localization and attribute correctness, addressing a limitation of standard Chamfer-distance mAP which evaluates geometry in isolation and ignores the semantic attributes required by planners. We further establish baselines on HRDX by benchmarking several representative methods for online vectorized HD map construction.

Our contributions are two-fold:
\begin{itemize}
    \item We release HRDX, the largest public dataset for HD map learning, featuring multi-view cameras, precise localization, LiDAR, aligned aerial imagery, and attribute-dense vector labels. Extensive experiments show that scale is a key driver of performance, consistently improving geometric fidelity and semantic attribute prediction.
    \item We demonstrate that distilling knowledge from aerial-augmented teachers significantly improves camera-only HD map construction while preserving deployment scalability.
\end{itemize}

\section{Related Work}
\subsection{Vector Map Datasets}
\label{sec:vector-map-datasets}
Several autonomous‑driving datasets provide vectorized high‑definition (HD) maps with structured polylines/polygons for lanes and boundaries, along with topological relations and traffic elements and we compare these datasets in Table \ref{tab:ds-compare-merged}. In summary, our dataset is over six times larger than previous largest HD-map dataset and provides a richer set of semantic layers and attributes.

\begin{table*}[t]
\centering
\begin{threeparttable}
\setlength{\tabcolsep}{4pt}
\renewcommand{\arraystretch}{1.1}
\scriptsize  
\begin{tabularx}{\textwidth}{@{}lccccc>{\raggedright\arraybackslash}X@{}}
\toprule
Dataset & \# Classes & \# Attr. & \# Hours & Aerial Imgs & Intersec Centerline &
Classes, Attributes present \\
\midrule
nuScenes \cite{caesar2020nuscenes} & 11 & 0 & 5   & \ding{55}* & \ding{55} &
lane, lane connector, drivable area, walkway, pedestrian crossing,
stop line, carpark area, traffic light. \\

Argoverse v1 \cite{chang2019argoverse} & 1 & 0 & 1.5  & \ding{55} & \ding{51} &
Lane centerlines with attributes (direction, intersection flags,
\texttt{has\_traffic\_control}), drivable area, and ground height. \\

Waymo Percep. \cite{sun2020scalability} & 7 & 7 & 6.4 & \ding{55} & \ding{51} &
Lanes (centerlines/boundaries), lane neighbors, road boundaries,
crosswalks, speed bumps, stop signs, traffic lights, driveway
entrances; features represented as 3D polylines/polygons. \\

Argoverse v2 \cite{wilson2023argoverse} & 4 & 0 & 4.1 & \ding{55} & \ding{51} &
3D lane boundaries (with marking types), vectorized drivable areas,
pedestrian crossings; lane attributes and topology. \\

\midrule
\textbf{HRDX (Ours)} & 10 & 20+ & 40+ & \ding{51} & \ding{51} &
Refer to Table~\ref{tab:annotation-ontology} for full ontology. \\
\bottomrule
\end{tabularx}
\begin{tablenotes}
\footnotesize
\item[*] {\scriptsize Some works augment nuScenes with aerial imagery obtained via Google Maps APIs for experimental use; however, systematic downloading or redistribution of such imagery is prohibited under Google’s Terms of Service.}
\item{\scriptsize Note: Waymo Open Motion Dataset \cite{ettinger2021large} is omitted from this table as it lacks full multi-camera image streams tied to map labels and is heavily concentrated on repeated routes.}
\end{tablenotes}
\end{threeparttable}
\caption{Comparison of public HD map datasets with HRDX.}
\label{tab:ds-compare-merged}
\end{table*}

\subsection{Automatic Vector Map Construction methods}

Recent Bird’s Eye View (BEV) perception approaches \cite{chen2024maptracker,wang2024stream} generate vectorized map elements (polylines) rather than relying on segmentation-based representations \cite{roddick2020predicting,dwivedi2021bird}. VectorMapNet \cite{liu2023vectormapnet} was the first end-to-end framework for polyline decoding, highlighting the limitations of rasterized map representations. Building on this, MapTR \cite{liao2022maptr,liao2025maptrv2} improves decoding stability through hierarchical queries and auxiliary tasks. StreamMapNet \cite{yuan2024streammapnet} introduces multi-point attention to better model polyline geometry, while MapTracker \cite{chen2024maptracker} further enhances temporal consistency via query propagation across frames.

\subsection{Aerial Imagery and Knowledge Distillation}

Aerial and satellite imagery have been widely used for remote sensing tasks and for extracting vectorized road or building topology \cite{yuan2021review,lv2023deep,shoaib2025advancements,gui2024remote,hua2025survey,he2022lane,li2019polymapper,vanetten2018spacenet}. In autonomous driving, aerial context has been shown to improve camera-based HD map construction by fusing satellite imagery with onboard sensors during training and inference \cite{gao2024complementing,lengerer2025aid4ad}. Cross-modal knowledge distillation (KD) has also been used to transfer LiDAR or fusion teachers to camera-only students. MapDistill \cite{hao2024mapdistill} applies this idea to HD map construction using a LiDAR–camera teacher.

\section{HRDX Dataset}
\label{sec:dataset}

In this section, we describe the setup of our data collection platform, the geographical scope of the drives, sensor synchronization, and the annotation protocol used to construct the dataset. 


\subsection{Vehicle Platform/  Sensor Suite}
\begin{figure}[htbp]
    \centering
    \includegraphics[width=0.9\linewidth]{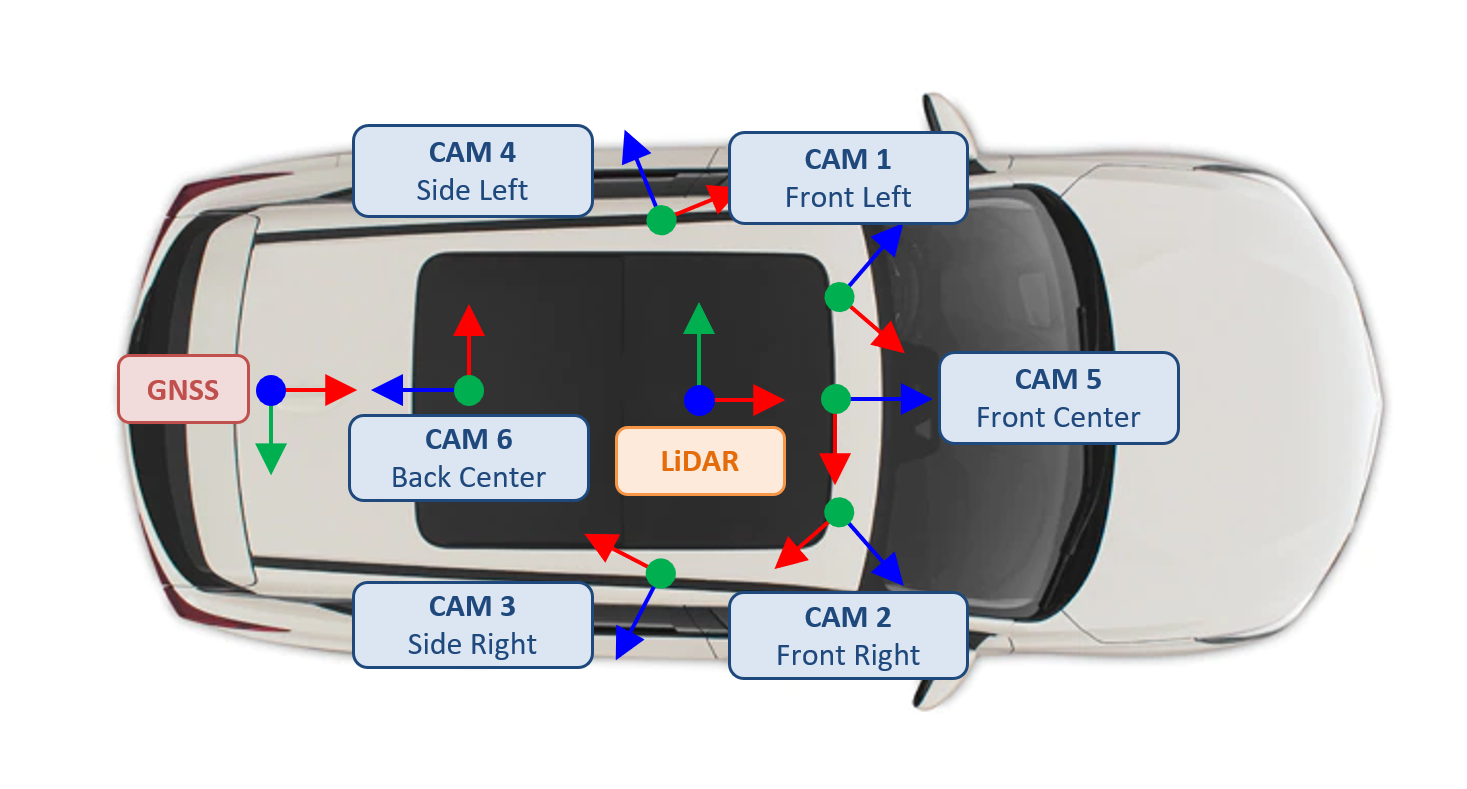} 
    \caption{Sensor setup of the data collection vehicle.}
    \label{fig:sensor_setup}
\end{figure}

Our data collection vehicle is equipped with a comprehensive multi-sensor suite designed for autonomous driving research. The platform integrates six wide-angle cameras, a high-resolution LiDAR, and a high-precision real-time kinematic (RTK) GNSS/INS system. All sensors are cross-calibrated and time-synchronized. A schematic top-view of the sensor placement is illustrated in Fig.~\ref{fig:sensor_setup}.

\subsection{Geographical Coverage}
\begin{figure*}[htbp]
    \centering
    \includegraphics[width=0.8\linewidth]{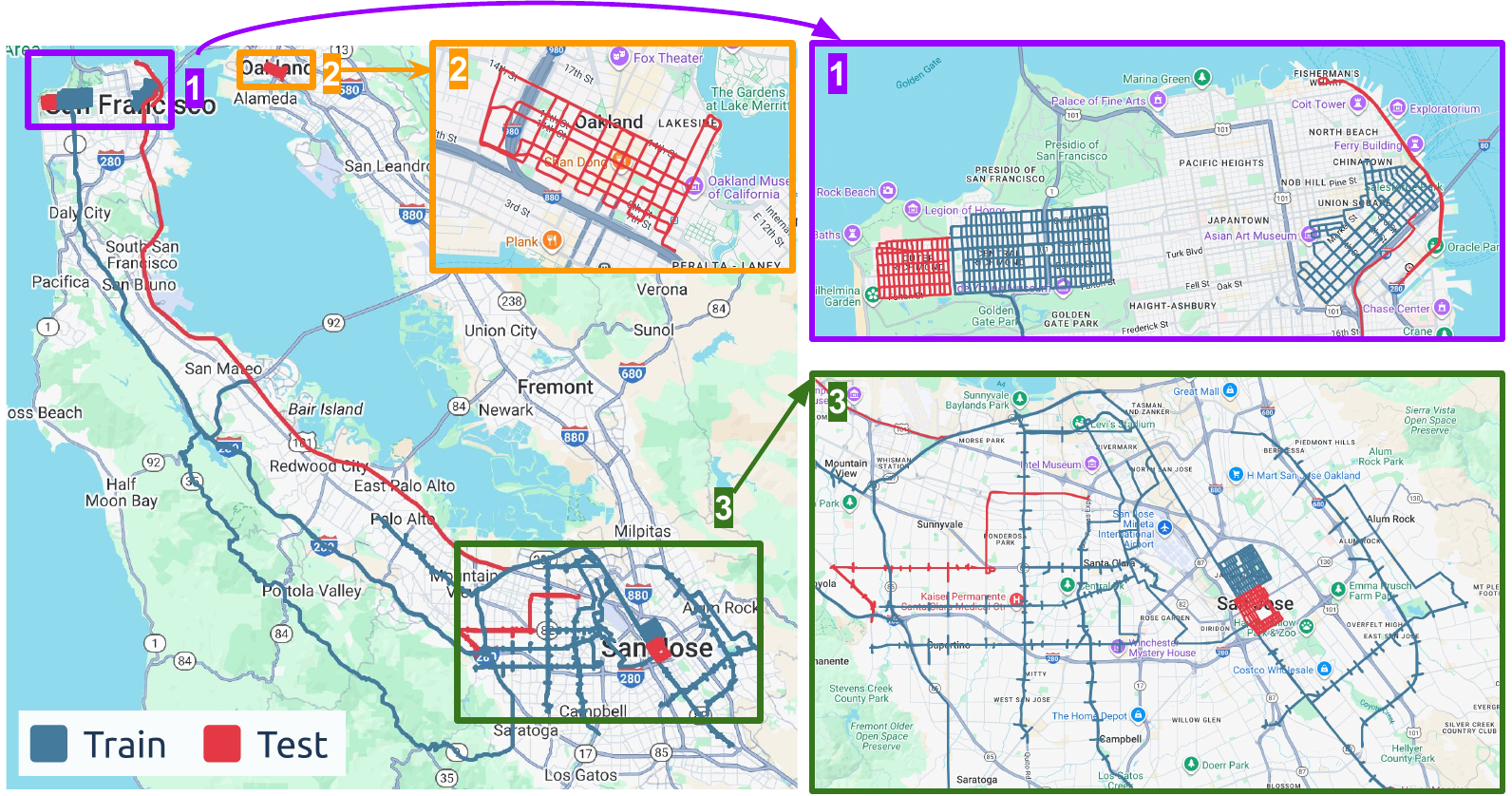} 
    \caption{Geographical distribution of our dataset along with the train and test splits}
    \label{fig:geographic_coverage}
\end{figure*}

Data collection was carried out in diverse driving environments in the San Francisco Bay Area. This includes dense urban centers, suburban regions, as well as major freeways. By encompassing both structured highway environments and complex urban settings characterized by dense traffic, frequent interactions, and visual occlusions, the dataset captures a wide range of real-world driving scenarios. Fig.~\ref{fig:geographic_coverage} illustrates the distribution of drive routes and our train / test splits across the region.

\subsection{Localization and Sensor Synchronization}
\begin{figure}[htbp]
    \centering
    \includegraphics[width=0.95\linewidth]{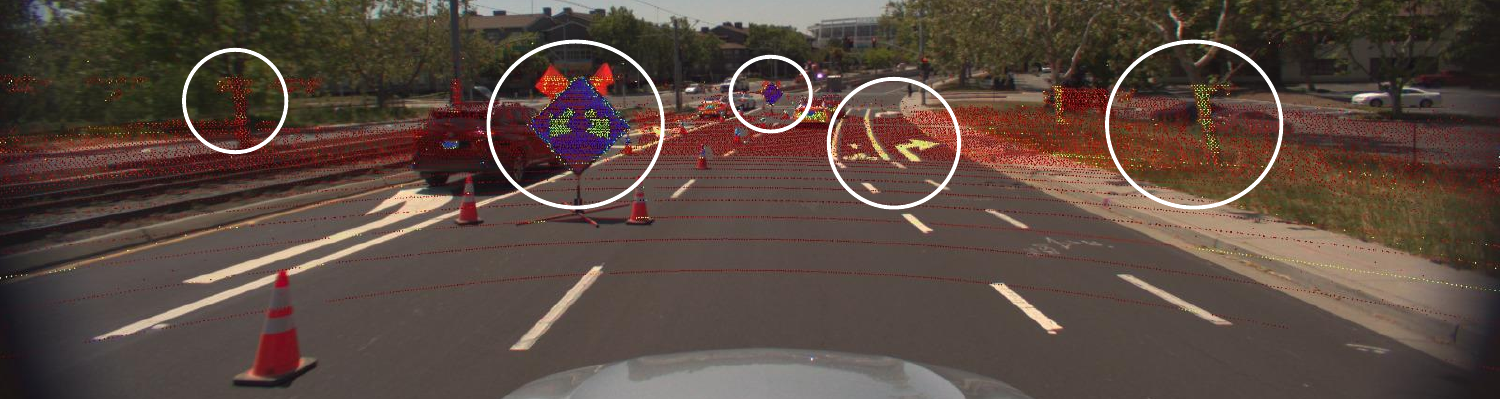} 
    \caption{Projection of point-cloud on image.}
    \label{fig:lidar_camera_porjection}
\end{figure}

High-precision INS (Inertial and Navigation System) using GNSS (Global Navigation Satellite System) with Real-Time Kinematic (RTK) correction is employed to obtain the ground-truth trajectory of the ego vehicle. In dense urban areas, GNSS degrades due to multipath and occlusions. We mitigate this by integrating LiDAR–Inertial Odometry (LIO) and fusing it with INS, yielding robust, drift-free localization. Ground truth includes geodetic position (latitude, longitude, altitude) and orientation, and full 6-DoF vehicle pose with respect to the first frame of each scene.\\
The GNSS/INS serves as the master clock. The six cameras operate as PTP (Precision Time Protocol) slaves to the GNSS/INS, while the LiDAR is triggered using the INS-provided PPS (Pulse Per Second) signal. LiDAR scans are motion-compensated to align with the corresponding image timestamps, mitigating distortions caused by ego-motion. Since camera exposures are nearly instantaneous, the resulting dataset achieves precise cross-modality synchronization, enabling pixel-accurate fusion between vision and range-based modalities. Figure \ref{fig:lidar_camera_porjection} demonstrates the high-quality alignment of LiDAR points on camera images, highlighting our precise sensor synchronization and calibration quality.

\subsection{Aerial Imagery}
\begin{figure}[htbp]
    \centering
    \includegraphics[width=0.95\linewidth]{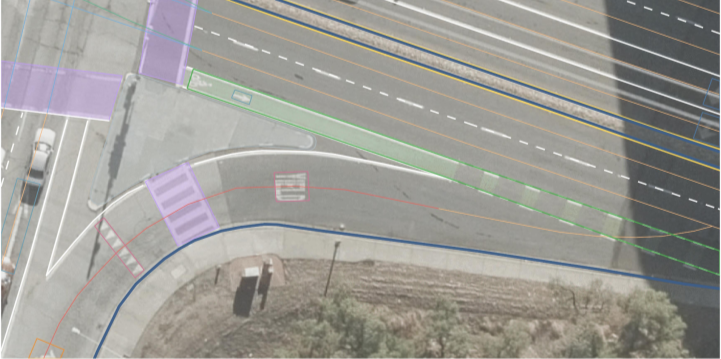} 
    \caption{Overlay of annotation on aerial imagery. (Refer to legend from Figure \ref{fig:teaser})}
    \label{fig:aerial overlay}
\end{figure}

To augment ground-level sensor data, we provide geo-referenced aerial imagery corresponding to each vehicle trajectory. Using poses derived from GNSS/INS, we retrieve high-resolution orthoimagery from publicly available county databases \cite{sf_aerial_2024,sanmateo_imagery_2022,sanjose_arcgis_nodate}. The imagery achieves a resolution of $8\,\mathrm{cm/pixel}$ in downtown areas and $15\,\mathrm{cm/pixel}$ in suburban and highway regions. This complementary overhead view facilitates research in map-based reasoning, top-down perception, and cross-view consistency, extending the utility of the dataset beyond conventional sensor modalities. Figure \ref{fig:aerial overlay} shows an aerial image overlaid with our HD map annotations, demonstrating precise spatial alignment.

\begin{figure*}[htbp]
    \centering
    \includegraphics[width=0.85\linewidth]{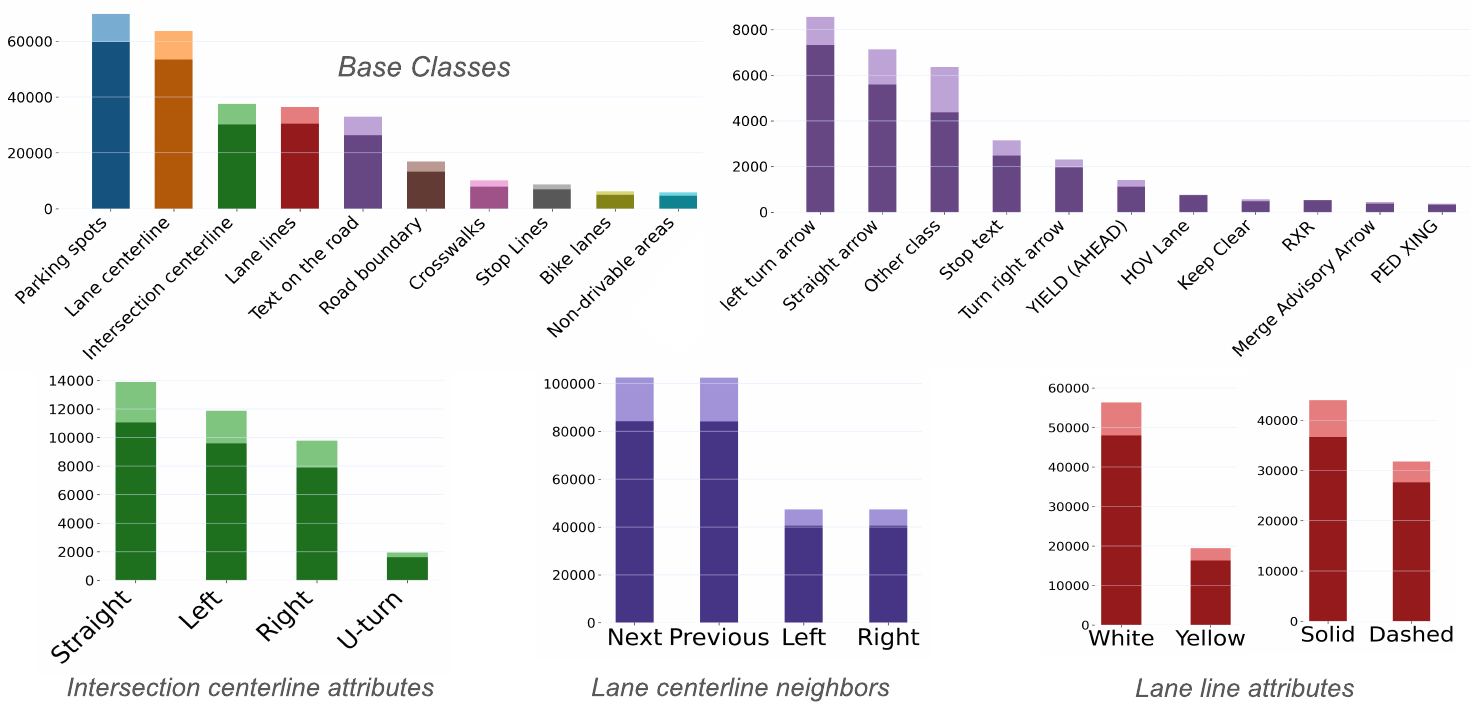} 
        \caption{Distribution of base classes and their attributes. For each bar, the darker segment represents items from the training split and the lighter segment represents items from the test split. See supplementary material for additional statistics.}
    \label{fig:dataset_hist}
\end{figure*}

\begin{table*}[htbp]
\centering
\scriptsize
\renewcommand{\arraystretch}{1.2}
\begin{tabular}{l l p{5.8cm} p{6.7cm}}
\toprule
\textbf{Class} & \textbf{Geometry} & \textbf{Attributes / Sub-classes} & \textbf{Description / Notes} \\
\midrule
Lane line & Polyline & Color (white / yellow), Type (solid / dashed) & Represents lane separators \\
Stop line & Polyline & None & Before intersections/crosswalks; marks vehicle stop area \\
Lane centerline & Polyline & Lane neighbors (previous / next / left / right) & Directional line defining drivable lane flow (not currently used in CS calculation) \\
Intersection centerline & Polyline & Turn type (left / straight / right / u-turn); multi-action possible & Defines allowed transitions between lanes at intersections \\
Crosswalk & Polygon & None & Pedestrian crossing region \\
Bike lane & Polygon & None & Dedicated bicycle path area \\
Parking spot & Polygon & None & Marked parking location \\
Road boundary & Polyline & None & Outer road edge separating roadway from sidewalk or terrain \\
Non-drivable area & Polygon & None & Surfaces where vehicles cannot drive (medians, curbs, islands) \\
\midrule
\makecell[l]{Text on road} & Polygon 
& Arrow (straight, left, right, left+straight, etc.), STOP, YIELD (AHEAD), HOV Lane, Railway Crossing (RXR), Pedestrian Crossing (PED XING), SCHOOL, SLOW, BUS / TAXI lane, Keep clear area, Other painted text 
& Each marking drawn as one polygon; grouped if multi-word \\
\bottomrule
\end{tabular}
\caption{Annotation ontology of the HRDX map dataset, summarizing geometric representation, attributes, and class-specific details.}
\label{tab:annotation-ontology}
\end{table*}


\subsection{Data Annotation and Ground-Truth Labels}
Our dataset comprises driving routes segmented into scenes. Each scene is a continuous sequence (on the order of hours) with time-synchronized multi-camera and LiDAR data; geo-referenced aerial imagery aligned with ego's pose. For every scene, we provide continuous HD vector maps. To construct reliable ground truth for static map elements, we accumulate LiDAR frames at an interval of 5m across each scene into a globally registered, high-density 3D map. This merged point cloud serves as the metric substrate for annotation. Expert annotators use the fused pointcloud to annotate road geometry in vector form (polylines and polygons) and encode topology (e.g., intersection centerlines), followed by visual verification to ensure spatial precision.

Static map features such as lane markings, road boundaries, crosswalks, and parking areas are labeled in vectorized form, following a standardized schema that encodes both geometry and semantics. Each element is represented using its native geometric primitive, \textit{polyline} for line-like entities and \textit{polygon} for area-like surfaces along with class-specific attributes capturing appearance or regulatory intent (e.g., color, dashed/solid type, turn allowance). 

A comprehensive summary of all annotated classes, their geometric types, associated attributes, and specific notes is provided in Table~\ref{tab:annotation-ontology}.  Figure \ref{fig:dataset_hist} shows the distribution of the classes and attributes in our dataset.

\section{Tasks and Metrics}

\label{hd-map-task}

The primary task in HRDX is online vectorized HD map construction. Given onboard sensor observations (e.g., surround-view cameras and LiDAR), the model predicts a local semantic map around the ego vehicle in vector form \cite{li2022hdmapnet, liu2023vectormapnet, liao2025maptrv2, liao2022maptr, chen2024maptracker}.

The output consists of sparse polylines or polygons representing map elements, jointly predicting geometry (lane centerlines, stop lines, road and lane boundaries, crosswalks) and attributes (lane color and style, text-on-road markings such as arrows, STOP, keep-clear, and intersection-related properties).

\subsection{Composite Evaluation Metric}
The HRDX benchmark includes ten semantic vector classes, several of which carry structured attributes (e.g., lane-line style, color etc). 
The conventional  Chamfer-distance mAP~\cite{li2022hdmapnet} evaluates only geometric localization of the classes, ignoring the correctness of the attributes. We therefore define a unified metric that combines geometric and attribute performance.

Following HDMapNet ~\cite{li2022hdmapnet}, we compute per-class average precision $\mathrm{AP}^{\text{geom}}_c$ under Chamfer-distance thresholds 
$\mathcal{T}=\{0.5,1.0,1.5\}\,\mathrm{m}$, averaged across thresholds:
\[
\mathrm{AP}^{\text{geom}}_c = \frac{1}{|\mathcal{T}|} \sum_{\tau \in \mathcal{T}} \mathrm{AP}^{\text{geom}}_c(\tau);
\mathrm{mAP}^{\text{geom}} = \frac{1}{|\mathcal{C}|}\sum_{c\in\mathcal{C}} \mathrm{AP}^{\text{geom}}_c
\]

For classes with attributes, we evaluate attribute prediction only on geometrically matched instances (at the mid threshold $\tau{=}1.0$\,m). For each attribute, we compute F$_1$ based on value equality between matched prediction–ground truth pairs. $\overline{\mathrm{F1}}^{\text{attr}}$ averages F$_1$ first across attributes within a class, then across attributed classes.

The composite score (CS) linearly combines geometry mAP and attribute F$_1$ using a dataset-level weight $\alpha$:
\[
\mathrm{Score} = (1-\alpha)\,\mathrm{mAP}^{\text{geom}} + \alpha\,\overline{\mathrm{F1}}^{\text{attr}},
\qquad
\alpha = \frac{|\mathcal{C}_{\text{attr}}|}{|\mathcal{C}|}
\]
Here, $\alpha$ is the fraction of classes with attributes, balancing geometric localization and attribute prediction in the overall score.

\section{Benchmark and Reference Baselines}
\label{Experiments}
In this section we benchmark online vector HD-map construction on HRDX using several camera-only baselines. We then select the best-performing model and study the impact of auxiliary modalities.  We denote the training/inference modality configuration as \textit{train/test}. We evaluate: (i) \textbf{C+L/C+L}, where camera and LiDAR is used during both training and inference; (ii) \textbf{C+A/C+A}, where camera and aerial imagery is used during both training and inference; and (iii) \textbf{C+A/C}, where camera and aerial imagery are used for training and inference uses cameras only.

\subsection{Camera-only Baselines}
We benchmark representative camera-only online vector HD-map models spanning diverse design families: StreamMapNet\cite{yuan2024streammapnet}, SQD-MapNet\cite{wang2024stream}, MapTRv2\cite{liao2025maptrv2}, and MapTracker\cite{chen2024maptracker}. All are trained with published schedules and a ResNet-50 backbone on HRDX. 
Results are summarized in Table~\ref{tab:map_element_comparison_baselines}. MapTracker achieves the highest mAP among camera-only models and is used as the base architecture for subsequent experiments.

\begin{table*}[t]
\centering
\footnotesize
\setlength{\tabcolsep}{2.5pt}
\renewcommand{\arraystretch}{1.2}
\resizebox{0.95\textwidth}{!}{
\begin{tabular}{l c c c c c c c c c c c c}
\hline
\textbf{Method} & \textbf{Mod.} &
\textbf{Lane} & \textbf{Stop} &
\textbf{Road} & \textbf{Lane} &
\textbf{Intersec.} & \textbf{Text} &
\textbf{NonDrv.} & \textbf{Park} &
\textbf{Cross} & \textbf{Bike} & \textbf{mAP} \\
 & \textbf{tr/te} &
\textbf{Line} & \textbf{Line} &
\textbf{Bdry} & \textbf{Ctr} &
\textbf{Ctr} & \textbf{Road} &
\textbf{Area} & \textbf{Spot} &
\textbf{walk} & \textbf{Lane} & \\
\hline
MapTRv2 \cite{liao2025maptrv2} & C/C & 56.85 & 30.75 & 61.26 & 64.73 & 33.96 & 35.18 & 14.03 & 12.09 & 43.82 & 38.52 & 39.12 \\
StreamMapNet \cite{yuan2024streammapnet} & C/C & 59.86 & 43.33 & 61.24 & 63.98 & 37.17 & 52.70 & 25.98 & 13.96 & 54.36 & 48.86 & 46.14 \\
SQDmapNet \cite{wang2024stream} & C/C & 57.69 & 37.11 & 60.25 & 62.14 & 35.87 & 47.62 & 22.39 & 12.07 & 51.49 & 46.71 & 43.33 \\
MapTracker (MT) \cite{chen2024maptracker}& C/C & 62.90 & 50.18 & 66.69 & 67.28 & 40.90 & 61.16 & 36.06 & 16.51 & 65.40 & 57.43 & 52.45 \\
\hline
\end{tabular}}
\caption{Map element detection performance compared across methods. All models trained and evaluated using on-board camera only.}
\label{tab:map_element_comparison_baselines}
\end{table*}

\begin{figure*}[!t]
    \centering
    \includegraphics[width=0.95\linewidth]{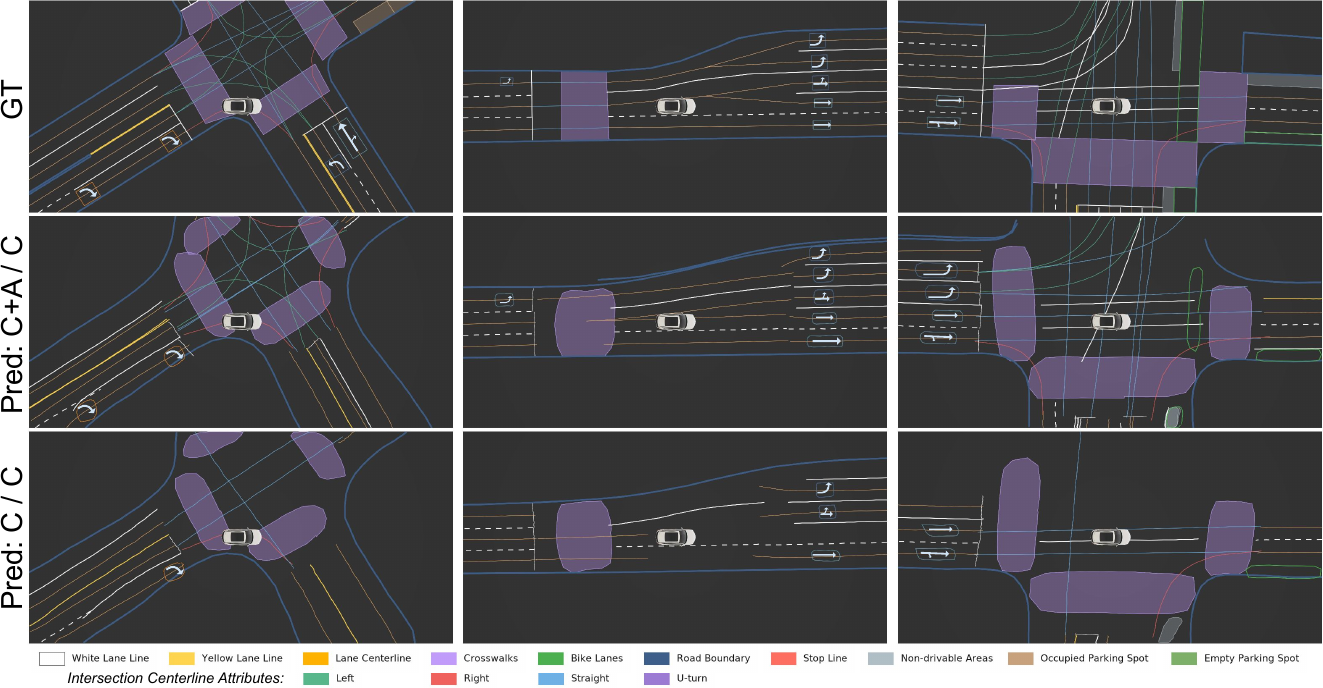} 
    \caption{Qualitative results on HRDX. The top row shows ground-truth annotations; the second row shows the camera-only model trained via aerial-based knowledge distillation (C+A / C in Table \ref{tab:map_element_comparison}); the third row shows the baseline camera-only model (C / C in Table \ref{tab:map_element_comparison}).}
    \label{fig:expt_results}
\end{figure*}

\begin{table*}[t]
\centering
\footnotesize
\setlength{\tabcolsep}{2.2pt}
\renewcommand{\arraystretch}{1.2}
\resizebox{0.95\textwidth}{!}{\begin{tabular}{l c c c c c c c c c c c c c c}
\hline
\textbf{Method} & \textbf{Mod.} &
\textbf{Lane} & \textbf{Stop} & \textbf{Road} & \textbf{Lane} &
\textbf{Intersec.} & \textbf{Text} & \textbf{NonDrv} & \textbf{Park} &
\textbf{Cross} & \textbf{Bike} & \textbf{mAP} & $\mathbf{\overline{\mathrm{F1}}^{attr}}$ & \textbf{CS} \\
 & \textbf{tr/te} &
\textbf{Line} & \textbf{Line} & \textbf{Bdry} & \textbf{Ctr} &
\textbf{Ctr} & \textbf{Road} & \textbf{Area} & \textbf{Spot} &
\textbf{walk} & \textbf{Lane} & & & \\
\hline
MT & C/C & 62.90 & 50.18 & 66.69 & 67.28 & 40.90 & 61.16 & 36.06 & 16.51 & 65.40 & 57.43 & 52.45 & 91.01 & 64.02 \\  
MT & C+A/C & 64.41 & 54.05 & 68.78 & 67.78 & 40.07 & 65.09 & 37.91 & 16.15 & 68.60 & 57.77 & 54.06 & 90.67 & 65.04 \\
MT & C+A/C+A & 62.95 & 62.29 & 72.56 & 66.40 & 41.82 & 67.18 & 36.79 & 18.81 & 73.29 & 56.26 & 55.86 & 89.99 & 66.09 \\
MT & C+L/C+L & 65.97 & 56.31 & 78.15 & 70.53 & 42.03 & 66.23 & 39.29 & 20.44 & 69.33 & 55.62 & 56.39 & 90.40 & 66.59 \\
\hline
\end{tabular}}
\caption{Comparison of different versions of MapTracker (MT). Modalities are C (camera), A (aerial imagery) and L (LiDAR). CS refers to our Composite Score.}
\label{tab:map_element_comparison}
\end{table*}

\subsection{Multimodal BEV Fusion} 
LiDAR is a standard auxiliary modality for geometric perception due to accurate depth measurements. In contrast, aerial imagery provides complementary global context: a top-down, occlusion-free view aligned to the BEV frame. This perspective encodes large-scale road topology, intersection layout, and surface markings that may be partially visible or occluded in egocentric views. We extend MapTracker to incorporate auxiliary BEV features from either LiDAR or aerial imagery. Each modality has a dedicated encoder (Camera BEV features are produced by the image backbone; ResNet‑50 U‑Net for aerial; SECOND‑style \cite{yan2018second} voxel encoder for LiDAR). Fusion is performed in BEV space using bidirectional cross-attention. Camera BEV features attend to auxiliary BEV features, followed by reverse attention from auxiliary to camera features. This ensures that the features of different modalities interact before fusion. The refined representations are concatenated and processed by a lightweight convolutional block to produce the fused BEV representation.

\subsection{Aerial-to-Camera Knowledge Distillation}
To transfer aerial context without increasing inference-time cost while avoiding aerial imagery at inference, we adopt a teacher–student framework with a camera+aerial teacher (C+A/C+A) and a camera-only MapTracker student. During training, the teacher produces predictions without gradient updates, while the student is optimized using supervised and distillation losses. Distillation is applied at the map head level, including classification distillation on teacher-positive queries, line regression distillation on geometry-matched pairs, and attribute distillation via KL divergence on logits. Teacher and student queries are matched using Hungarian assignment. 

\subsection{Attribute Heads}
We augment the map head with lightweight attribute classifiers. For each set of attributes, we attach a linear classifier on top of every decoder query embedding and use cross-entropy loss.
For classes with attributes, we measure per-attribute precision, recall and f-score computed on geometry-matched predictions only to avoid penalizing localization twice.
\begin{table}[htbp]
\centering
\small
\renewcommand{\arraystretch}{1.1}
\setlength{\tabcolsep}{3pt} 
\begin{adjustbox}{max width=\linewidth}
\begin{tabular}{r l r l}
\toprule
\textbf{Attribute} & \textbf{F$_1$ (P/R)} & \textbf{Attribute} & \textbf{F$_1$ (P/R)} \\
\midrule

\multicolumn{4}{@{}l@{}}{\textbf{Intersection centerline attributes} (\textit{mean F$_1$} = \textbf{0.862})} \\
straight & 0.895 (0.94/0.86) & left   & 0.864 (0.83/0.91) \\
right    & 0.899 (0.89/0.91) & u-turn & 0.789 (0.76/0.82) \\
\midrule

\multicolumn{4}{@{}l@{}}{\textbf{Lane color} (\textit{mean F$_1$} = \textbf{0.975})} \\
white  & 0.985 (0.99/0.98) & yellow & 0.964 (0.96/0.97) \\
\midrule

\multicolumn{4}{@{}l@{}}{\textbf{Lane style} (\textit{mean F$_1$} = \textbf{0.957})} \\
solid  & 0.965 (0.95/0.98) & dashed & 0.949 (0.97/0.93) \\
\midrule

\multicolumn{4}{@{}l@{}}{\textbf{Text on road} (\textit{mean F$_1$} = \textbf{0.795})} \\
left-turn arrow & 0.914 (0.90/0.93) & straight arrow & 0.820 (0.89/0.76) \\
ped xing        & 0.503 (0.52/0.49) & merge advisory & 0.685 (0.59/0.82) \\
yield ahead     & 0.803 (0.84/0.77) & stop text      & 0.949 (0.96/0.94) \\
other           & 0.891 (0.87/0.91) &                &                  \\
\bottomrule
\end{tabular}
\end{adjustbox}
\caption{Per-attribute performance (F$_1$ score, with Precision (P) / Recall (R) in parentheses) for MapTracker (C/C).}
\label{tab:attr-perf}
\end{table}

\subsection{Implementation Details} 
All experiments are trained on 4$\times$H100 GPUs for 20 epochs. For MapTracker, we follow the original three-stage schedule (6, 2, and 12 epochs). To reduce computation without losing road cues, we crop the top 36\% of each camera image (mostly sky) at train/test and update the intrinsics accordingly. All experiments use a 60m$\times$30m BEV region centered on the ego vehicle.

\section{Analysis}
\subsection{Dataset Scale}
\begin{figure}[t]
    \centering
    \includegraphics[width=\linewidth]{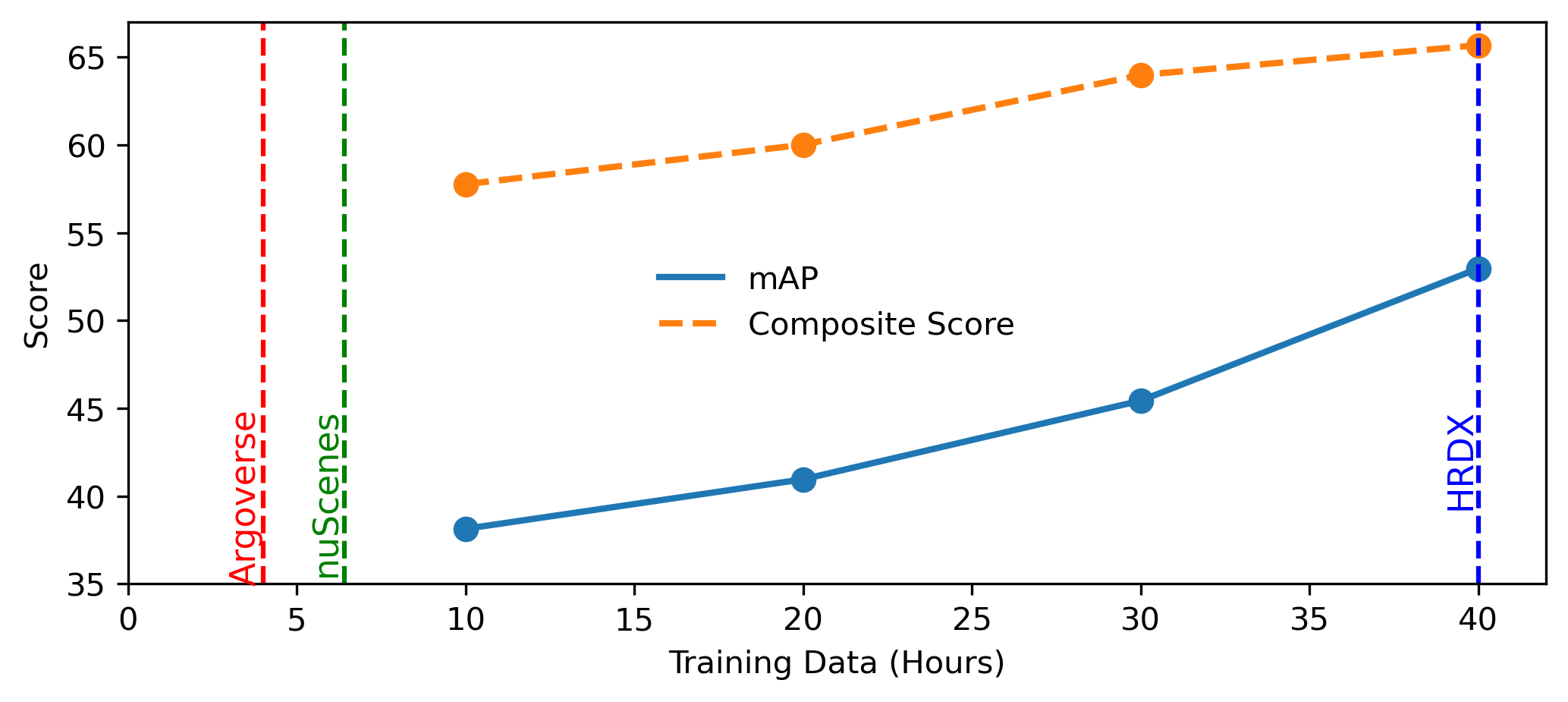} 
    \caption{Composite score and mAP of Maptracker(C/C) is plotted against the fraction of dataset the model is trained on. Scales of Argoverse and Nuscenes are denoted by vertical lines.}
    \label{fig:dataset_size}
\end{figure}

A key contribution of HRDX is its scale in both driving coverage and temporal extent. For context, nuScenes is curated as 1{,}000 scenes of fixed 20\,s duration \cite{caesar2020nuscenes}. In contrast, HRDX provides long, contiguous temporal sequences spanning hours, enabling evaluation of online mapping under sustained temporal propagation (including drift and error accumulation). 
To quantify the impact of dataset scale, we train MapTracker (C/C) on 25\%, 50\%, and 75\% of the HRDX dataset, keeping the same test sets. As shown in Figure~\ref{fig:dataset_size}, both mAP and CS improve monotonically with data scale, indicating improved learning stability and generalization. These results suggest that modern temporal mappers benefit from large and diverse driving coverage.
\begin{figure*}[!t]
\centering
\includegraphics[width=0.85\linewidth]{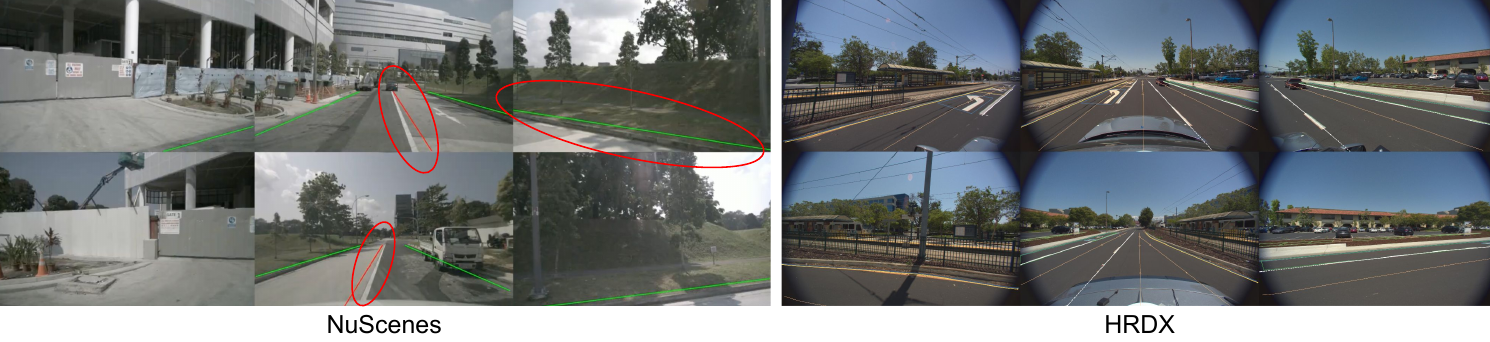}
\caption{Annotation quality: HD map polylines projected onto camera images for nuScenes and HRDX illustrating HRDX's superior quality. In nuScenes, green denotes curb polylines. The HRDX legend follows Figure \ref{fig:teaser}.}
\label{fig:quality_fig}
\end{figure*}
\subsection{Rich attributes and high-fidelity annotation}
\label{sec:rich-attrs}
HD maps consumed by downstream planners encode regulatory and behavioral semantics such as lane color and style (for legal lane changes), stop lines and road text (for rule-compliant stopping), intersection maneuver constraints, and dedicated lanes such as bike or HOV, that cannot be recovered from geometry alone~\cite{bao2023hdmapreview}. Prior public HD-map datasets leave most of these semantics unlabeled, as shown in Table~\ref{tab:ds-compare-merged}. HRDX annotates 10 base classes and 20+ attributes (Table~\ref{tab:annotation-ontology}); several of which are unique to HRDX and exhaustively cover all the static road elements for planning.

This richer supervision is learnable but far from saturated. As shown in Table~\ref{tab:attr-perf}, common attributes like lane color (0.975) and style (0.957) approach ceiling performance, while rarer cases such as u-turns (0.789) and text-on-road categories (e.g., pedestrian crossing 0.503, merge advisory 0.685) remain challenging. These difficult attributes are absent in prior HD-map datasets yet capture critical regulatory semantics for planning.

Annotation fidelity matters for both geometry and attribute learning. By labeling on a globally registered point cloud accumulated at 5,m intervals, HRDX polylines align tightly with visible markings and curbs, with reduced offsets compared to nuScenes~\cite{caesar2020nuscenes} (Fig.~\ref{fig:quality_fig}; more in supplementary).

\subsection{Leveraging aerial imagery}
\label{sec:analysis_aerial}
Aerial orthoimagery is a strong \emph{scaling lever} for HD-map learning: it provides a globally consistent top-down view with wide coverage and minimal ego-occlusion, and can be obtained without specialized mapping fleets.  HRDX is the first HD-map dataset to include legally redistributable aerial imagery aligned to the BEV target frame.

Adding aerial imagery during training and inference (MT, C+A/C+A) improves the camera-only baseline by +3.41 mAP (52.45 $\rightarrow$ 55.86) and +2.07 CS (64.02 $\rightarrow$ 66.09) (Table~\ref{tab:map_element_comparison}). The gains concentrate on elements that benefit from global layout and reduced occlusion: stop lines improve by +12.11, road boundaries by +5.87, crosswalks by +7.89, and text-on-road by +6.02. These are precisely the classes that are expensive to annotate exhaustively from egocentric views due to viewpoint variability, truncation, and occlusion.

Aerial imagery is typically not available to an onboard system at inference, but it is readily available offline during training. Our results show that this offline signal can still improve a camera-only mapper: using aerial imagery only during training (MT, C+A/C) yields +1.61 mAP (52.45 $\rightarrow$ 54.06) and +1.02 CS (64.02 $\rightarrow$ 65.04) over the camera-only baseline (Table~\ref{tab:map_element_comparison}). The use of aerial imagery during training closes the gap between camera-only inference and camera+lidar inference. This supports the premise that aerial imagery can function as a privileged, scalable prior whose benefits can be distilled into a deployment-feasible camera-only model.

\section{Conclusion}
\label{Conclusion}
We introduce HRDX, the largest publicly available dataset for online vector HD-map construction, with rich semantic annotations, multi-sensor data, and aligned aerial imagery. We further demonstrate that aerial imagery boosts mapping both as a direct input and as a training-only signal via knowledge distillation.

{
    \small
    \bibliographystyle{ieeenat_fullname}
    \bibliography{main}
}

\end{document}